\documentclass[10pt,letterpaper]{article}
\usepackage[top=0.85in,left=2.75in,footskip=0.75in,marginparwidth=2in]{geometry}

\usepackage[utf8]{inputenc}

\usepackage{cite}

\usepackage{nameref,hyperref}

\usepackage[right]{lineno}

\usepackage{microtype}
\DisableLigatures[f]{encoding = *, family = * }

\raggedright
\usepackage{changepage}

\usepackage[aboveskip=1pt,labelfont=bf,labelsep=period,singlelinecheck=off]{caption}

\makeatletter
\renewcommand{\@biblabel}[1]{\quad#1.}
\makeatother

\usepackage{lastpage,fancyhdr,graphicx}
\usepackage{epstopdf}
\usepackage{natbib}
\usepackage{filecontents}
\fancyheadoffset[L]{2.25in}
\fancyfootoffset[L]{2.25in}

\usepackage{color}

\definecolor{Gray}{gray}{.25}

\usepackage{mathtools, nccmath}
\usepackage{mathrsfs}
\usepackage{mathptmx}
\usepackage{wrapfig}
\usepackage[pscoord]{eso-pic}
\usepackage[fulladjust]{marginnote}
    
\usepackage{url}
\usepackage{xcolor}
\definecolor{newcolor}{rgb}{.8,.349,.1}
\usepackage{mathtools, nccmath}
\usepackage{multicol, blindtext}
\usepackage{graphicx}
\usepackage{mathrsfs}
\usepackage{rotating}
\usepackage{blindtext}
\usepackage{mathptmx}
\usepackage{anyfontsize}
\usepackage{t1enc}
\usepackage{amsmath, amssymb, txfonts}
\usepackage{amsmath}
\usepackage{rotating}
\usepackage{lipsum}
\usepackage{graphicx}
\usepackage{placeins}

\usepackage{amssymb}
\usepackage{latexsym}
\usepackage{mathtools}
\usepackage{url}
\usepackage{xcolor}
\usepackage{graphicx}
\usepackage{multicol}

\begin{document}
\vspace*{0.35in}

\begin{flushleft}
{\Large
\textbf\newline{Weighted second-order cone programming twin support vector machine for imbalanced data classification}
}
\newline
\\
Saeideh Roshanfekr\textsuperscript{1},
Shahriar Esmaeili\textsuperscript{2}
Hassan Ataeian\textsuperscript{3}, and
Ali Amiri\textsuperscript{3}
\\
\bigskip
\bf{1} {Department of Computer Engineering and Information Technology, Amirkabir University of Technology, 424 Hafez Avenue, 15875-4413 Tehran, Iran}
\\
\bf{2} {Department of Physics and Astronomy, Texas A$\&$M University, 4242 TAMU, University Dr., College Station, TX 77840, US}
\\
\bf{3} {Department of Computer Engineering, University of Zanjan, University Blvd., 45371-38791 Zanjan, Iran}

\bigskip

\end{flushleft}

\begin{abstract}
We propose a method of using a Weighted second-order cone programming twin support
vector machine (WSOCP-TWSVM) for imbalanced data classification. This method
constructs a graph based under-sampling method which is utilized to remove outliers and
reduce the dispensable majority samples. Then, appropriate weights are set in order to decrease the impact of samples of the majority class and increase the effect of the minority class
in the optimization formula of the classifier. These weights are embedded in the optimization
problem of the Second Order Cone Programming (SOCP) Twin Support Vector Machine
formulations. This method is tested, and its performance is compared to previous methods on
standard datasets. Results of experiments confirm the feasibility and efficiency of the
proposed method.

\end{abstract}

\section{Introduction}
\label{sec1}
The imbalanced problem for classification methods is the basic issue of research in data mining. Datasets are said to be imbalanced if the samples belonging to majority class outnumbers the data samples belonging to the minority class. Many attempts have been made to deal with this problem in various context, such as credit scoring \citep{Brown_2012}, fraud detection \citep{Phua_2010}, spam filtering \citep{Tang_2006} and anomaly detection \citep{Pichara_2011}. When the training dataset is imbalanced, the difference between the performance of the majority class and minority class becomes larger. To solve this problem, two methods have been proposed: one is based on the sampling method and the other one is a cost-sensitive method. Sampling method can be divided into two classes: under-sampling method and over-sampling method. In the under-sampling, the training dataset is reduced in the majority training set so this this dataset is balanced samples of the majority sets. In the over-sampling method, data from the minority class are copied multiple times or slightly changed such that the two classes are balanced. Many issues have been made in this context, such as Random under-sampling, Random over-sampling \citep{Kotsiantis_2006}, SMOTE \citep{Chawla_2011}, MSMOTE \citep{Phua_2010}, Random Walk over-sampling \citep{Zhang_2014}. An hybrid method that selects features in high dimensional datasets has also been proposed by \cite{Moradkhani_2015}. Recently \cite{ataeian_2019} investigated a method for large margin classifiers. 

The second approach to the imbalanced data classification problem is to apply the weights of the training data points \citep{Elkan_2001, Zadrozny_2003, Zhi_Hua_Zhou_2006}. Twin SVM is one of the extensions of SVM which constructing two classifiers in such a way that each one is close to one of the two classes. Note that for the imbalanced problem, the standard SVM has been modified by many researchers \citep{Suykens_2002, Deng_2012, Tomar_2014, Shao_2014}. The Second-Order Cone Programming (SOCP) formulations have been proposed for SVM and Twin SVM. These formulations consider all possible choices of class-conditional densities in a way that with a given mean and covariance matrix and also with having two constraints, one for each class results in a much more efficient training \citep{Nath_2007, Maldonado_2016}. SOCP-TWSVM constructs two nonparallel classifiers in a way that each hyperplane is closer to one of the training patterns and at the same time as far as possible from the other. Each training pattern is represented by an ellipsoid characterized by the mean and covariance of each class.

In this paper, we attempt to extend the SOCP-TWSVM of imbalanced datasets. The proposed Weighted SOCP-TWSVM (WSOCP-TWSVM) has two phases. Firstly, it utilizes a graph-based under-sampling method to remove outliers and reduce the dispensable majority samples. Then, a weighted bias is introduced to decrease the impact of samples of the majority class and increase the effect of minority class in the optimization formula of the classifier. The SOCP is utilized to solve the model. The methods Twin-SVM, SOCP-TWSVM for binary classification are introduced in Section \ref{sec2}. The proposed approach is discussed in Section \ref{sec3}. The Experimental results are given in Section \ref{sec4}. The main conclusions and future works have also been provided in Section \ref{sec5}.

\section{Preliminaries}\label{sec2}

\subsection{Twin SVM}
Twin SVM \citep{Khemchandani_2007} is a classification method which separates the instances by constructing two nonparallel hyperplanes instead of a hyperplane. The hyperplanes are obtained by solving two small size optimization problem using QPP. The parameters of the hyperplanes are calculated by solving the following optimization problems:
\begin{footnotesize}
\begin{equation}\label{eq1}
\begin{matrix}
  \begin{matrix}
    min \\
    W_{1}, b_{1},\xi_{2}
  \end{matrix} & \frac{1}{2}||AW_{1}+e_{1}b_{1}||^2 + \frac{c_{1}}{2}(||W_{1}||^2+b_{1}^2)+c_{3}e_{2}^T\xi_{2} \\
   & \\
  subject~to & - (BW_{1}+e_{2}b_{1})\geq e_{2} - \xi_{2},\quad \xi_{2}\geq 0,
\end{matrix}
\end{equation}
\end{footnotesize}
\begin{footnotesize}
\begin{equation}\label{eq2}
\begin{matrix}
  \begin{matrix}
    min \\
    W_{2}, b_{2},\xi_{1}
  \end{matrix} & \frac{1}{2}||BW_{2}+e_{2}b_{2}||^2 + \frac{c_{2}}{2}(||W_{2}||^2+b_{2}^2)+c_{4}e_{1}^T\xi_{1} \\
   & \\
  subject~to & - (AW_{2}+e_{1}b_{2})\geq e_{1} - \xi_{1},\quad \xi_{1}\geq 0,
\end{matrix}
\end{equation}
\end{footnotesize}
where $c_{1}, c_{2}, c_{3}, c_{4}$ are positive parameters, and $e_{1}$ and $e_{2}$ are vectors of one of appropriates dimensions.  Parameters $c_{3}$ and $c_{4}$ determine the trade-off between the respective model fit and the summation of the slack variables.

\subsection{SOCP-TWSVM}
This classifier has combined the ideas of Twin SVM and SOCP-SVM. The reasoning behind this approach is developing two nonparallel classifiers in a way that each hyperplane is closest to one of the two classes and also in the same distance from the other class, \citep{Maldonado_2016}. This problem can be formulated as the following quadratic programming model

\begin{footnotesize}
\begin{equation}\label{eq3}
\begin{matrix}
  \begin{matrix}
    min \\
    W_{1}, b_{1}
  \end{matrix} & \frac{1}{2}||AW_{1}+e_{1}b_{1}||^2 + \frac{\theta_{1}}{2}(||W_{1}||^2+b_{1}^2) \\
   & \\
  subject~to & - (W_{1}^{T}\mu_{2}-b_{1})\geq 1 + \kappa_{2}||S_{2}^T W_{1}||,
\end{matrix}
\end{equation}
\end{footnotesize}

\begin{footnotesize}
\begin{equation}\label{eq4}
\begin{matrix}
  \begin{matrix}
    min \\
    W_{2}, b_{2}
  \end{matrix} & \frac{1}{2}||BW_{2}+e_{2}b_{2}||^2 + \frac{\theta_{2}}{2}(||W_{2}||^2+b_{2}^2) \\
   & \\
  subject~to & - (W_{2}^{T}\mu_{1}-b_{2})\geq 1 + \kappa_{1}||S_{1}^T W_{2}||,
\end{matrix}
\end{equation}
\end{footnotesize}

where $\theta_{1}, \theta_{2}>0$, $\Sigma_{i} = S_{i}S_{i}^T$ and $\Sigma_{i}$  is covariance of each class, and $\kappa_{i}=\sqrt{\frac{\eta_{i}}{1-\eta_{i}}}$ which defined as probability of false- negative (false-positive) errors.

\section{WSOCP-TWSVM: Weighted SOCP-TWSVM}\label{sec3}
In this section, we present the proposed Weighted Second-Order Cone Programming Twin Support Vector Machine (WSOCP-TWSVM) for the imbalanced problem. This classifier removes outliers and reduces the unessential majority samples with a graph-based under-sampling method. Also, a weighted bias is presented to control the impact of the samples of each class. These weights define the sensitivity of the classifiers to the imbalance ratio and are considered in the mathematical model of the classifier.

\subsection{Sampling method}\label{subsec3.1}
\begin{figure}[ht]
\centerline{\includegraphics[width=1\textwidth,clip=]{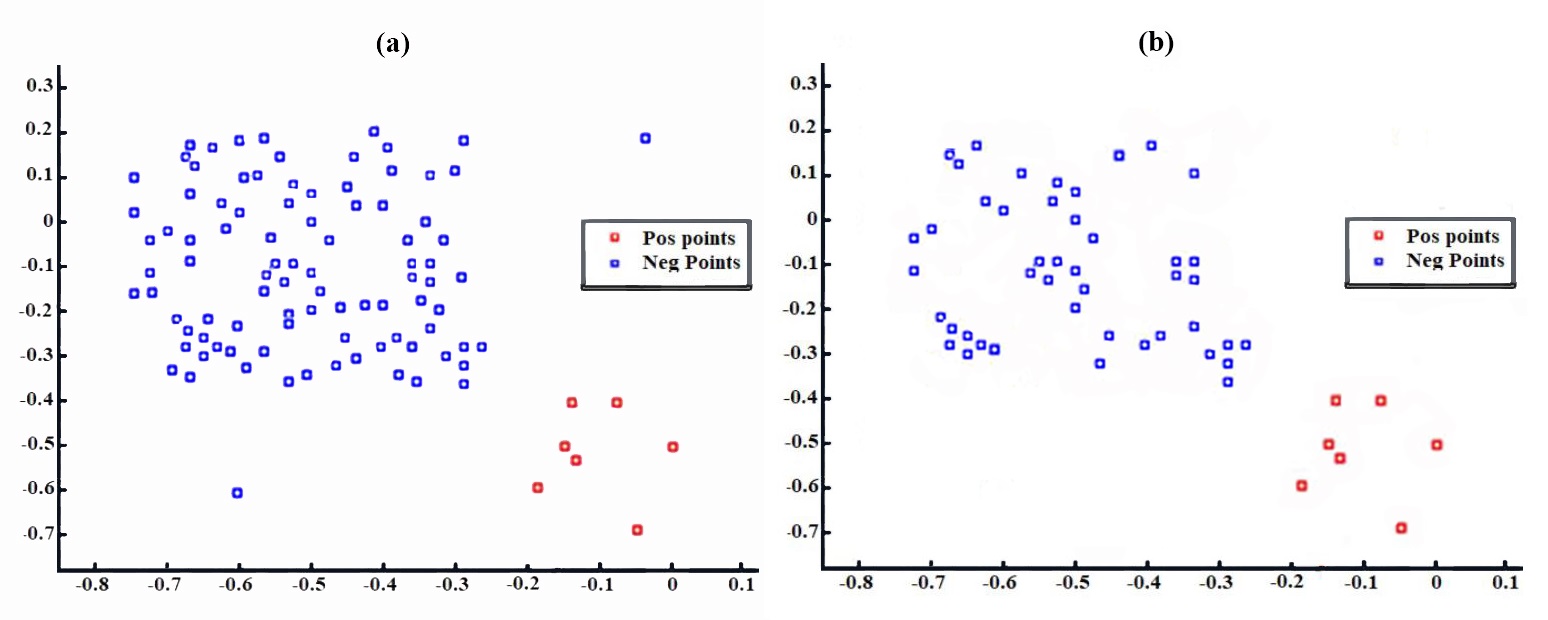}}
\caption{Decreasing the number of majority class, (a) before sampling, (b) after sampling. }\label{fig1}      
\end{figure}

In this method, supposing that the samples of minority class remain unchanged, and the samples of majority class are selected by constructing a proximity graph \citep{Belkin_2006, Yang_2009}. The samples with nonzero degree are in high density regions; and the samples with zero degree such as outliers are in low density regions. The adjacent matrix, $U$, is defined as
follows,
\begin{footnotesize}
\begin{equation}\label{eq7}
U_{ij} = \left\{
\begin{array}{rl}
\tau_{ij}, & \qquad x_{i} \in N_{k}(j)\quad and \quad x_{j} \in N_{k}(i) \\
0 & \qquad otherwise\\

\end{array} \right.
\end{equation}
\end{footnotesize}
where $(N_{k}(j))$ is a set of the k-nearest neighbors in the majority class of the point $x_{j}$, $(N_{k}(j))$ k-nearest neighbors of the point $x_{i}$, and $U$ is adjacent matrix. $\tau_{ij}$ is a a scalar value or any characters for showing k-neareast neighbors in a special vertex. $\tau_{ij}$ can be assumed any amount expect zero, and $i,j=1,..,n$. Then we define the under-sampling coefficient as
\begin{footnotesize}
\begin{equation}\label{eq8}
u_{i} = \left\{
\begin{array}{rl}
1, & \qquad \sum_{j} U_{ij}\geq k \\
0 & \qquad otherwise\\
\end{array} \right.
\end{equation}
\end{footnotesize}
where all points with nonzero $u_{i}$ are selected as members of the sample set, Fig. \ref{fig1}.

\subsection{Defining bias weights}\label{subsec3.2}
In imbalanced problems, setting of the appropriate weights to the samples of training set is a critical issue
in cost-sensitive approaches. The data in the majority class have to receive lower weight than those in the minority class. Also, the weight should be in (0,1) state. If the size of positive class is $N_{pos}$ and that of negative set after undersampling is $N_{-neg}$, the weights are defined as
\begin{footnotesize}
\begin{equation}\label{eq9}
D_{1} = \left\{
\begin{array}{rl}
1 & \qquad N_{pos} \geq N_{-neg} \\
\frac{N_{-neg}}{N_{pos}} & \qquad N_{pos} < N_{-neg}\\

\end{array} \right.
\end{equation}
\end{footnotesize}
\begin{footnotesize}
\begin{equation}\label{eq10}
D_{2} = \left\{
\begin{array}{rl}
\frac{N_{-neg}}{N_{pos}} & \qquad N_{pos} \geq N_{-neg} \\
 1& \qquad N_{pos} < N_{-neg}\\

\end{array} \right.
\end{equation}
\end{footnotesize}
where $N_{-neg}$ and $N_{pos}$ are the size of negative and positive classes after sampling.
\subsection{Linear Weighted SOCP twin SVM}\label{subsec3.3}
WSOCP-TWSVM combines the graph-based under-sampling and the previous weighting methods.
First, performing the under-sampling method described in Subsection \ref{subsec3.1}, discards instances from the
majority class and the remaining is demonstrated by $B_{\_}$. Then, the weight of the two classes will be
calculated using Eq. (\ref{eq9}) and Eq. (\ref{eq10}). The majority and minority hyperplanes are determined by
solving the following optimization equations

\begin{footnotesize}
\begin{equation}\label{eq11}
\begin{matrix}
  \begin{matrix}
    min \\
    W_{1}, b_{1}, \xi_{2}
  \end{matrix} & \frac{1}{2}||AW_{1}+e_{1}b_{1}||^2 + \frac{\theta_{1}}{2}(||W_{1}||^2+b_{1}^2)+c_{1}D_{1}\xi_{2} \\
   & \\
  subject~to & - (W_{1}^{T}\mu_{2}-b_{1})\geq 1 - \xi_{2} + \kappa_{2}||S_{2}^T W_{1}||.\xi_{2}\geq 0,
\end{matrix}
\end{equation}
\end{footnotesize}

\begin{footnotesize}
\begin{equation}\label{eq12}
\begin{matrix}
  \begin{matrix}
    min \\
    W_{2}, b_{2}, \xi_{1}
  \end{matrix} & \frac{1}{2}||B_{\_}W_{2}+e_{2}b_{2}||^2 + \frac{\theta_{2}}{2}(||W_{2}||^2+b_{2}^2)+c_{2}D_{2}\xi_{1} \\
   & \\
  subject~to & - (W_{2}^{T}\mu_{1}-b_{2})\geq 1 - \xi_{1} + \kappa_{1}||S_{1}^T W_{2}||.\xi_{1}\geq 0,
\end{matrix}
\end{equation}
\end{footnotesize}
where $\theta_{1}, \theta_{2}, c_{1}, c_{2}>0, \Sigma_{2-} = S_{2-}S^T_{2-}$, and $\xi_{i}$ are defined as slack variables- soft margin error of the $i-th$ training point, $\mu_{i}$ is mean of each class.

Lagrangian function under Karush-Kuhn-Tucker condition and associated with Eqs. (\ref{eq11}) and (\ref{eq12}) can also be rewritten as
\begin{footnotesize}
\begin{multline}\label{eq15}
\hat{\mathscr{L}}(W_{1}, b_{1}, \lambda_{1}, \rho_{1}, \xi_{2}, u_{2})=\frac{1}{2}||AW_{1}+e_{1}b_{1}||^2 + \frac{\theta_{1}}{2}(||W_{1}||^2+b_{1}^2)+ \\
 c_{1}D_{1}\xi_{2}= \lambda_{1}(W_{1}^T \mu_{2-}+b_{1}+1-\xi_{2}+\kappa_{2}W_{1}^T S_{2-}u_{2})-\rho_{1}\xi_{2}
\end{multline}
\end{footnotesize}

\begin{footnotesize}
\begin{multline}\label{eq16}
\hat{\mathscr{L}}(W_{2}, b_{2}, \lambda_{2}, \rho_{2}, \xi_{1}, u_{1})=\frac{1}{2}||B_{\-}W_{2}+e_{2}b_{2}||^2 + \frac{\theta_{1}}{2}(||W_{1}||^2+b_{1}^2)+ \\
 c_{2}D_{2}\xi_{1}= \lambda_{2}(-W_{2}^T \mu_{1}+b_{2}+1-\xi_{1}+\kappa_{2}W_{2}^T S_{1}u_{1})-\rho_{2}\xi_{1}
\end{multline}
\end{footnotesize}
Consequently, the dual problem can be stated as follows;
\begin{footnotesize}
\begin{equation}\
\begin{matrix}
  \begin{matrix}
    max \\
    z_{1}, u_{2}, \lambda_{1}
  \end{matrix} & \frac{1}{2}\lambda_{1}^2 Z_{1}^T(H^{T}H+\theta_{1}I)^{-1}Z_{1}+\lambda_{1} \\
   & \\
  subject~to & z_{1}=\mu_{2-}+\kappa_{2}S_{2-}u_{2},||u_{2}||\leq 1,\qquad 0\leq \lambda_{1}\leq c_{1}D_{1}, 
\end{matrix}
\end{equation}
\end{footnotesize}
\begin{footnotesize}
\begin{equation}\
\begin{matrix}
  \begin{matrix}
    max \\
    z_{2}, u_{1}, \lambda_{2}
  \end{matrix} & \frac{1}{2}\lambda_{2}^2 Z_{2}^T(G^{T}G+\theta_{2}I)^{-1}Z_{2}+\lambda_{2} \\
   & \\
  subject~to & z_{2}=\mu_{1}+\kappa_{1}S_{1}u_{1},||u_{1}||\leq 1,\qquad 0\leq \lambda_{2}\leq c_{2}D_{2},
\end{matrix}
\end{equation}
\end{footnotesize}
where, $H = [A, e_1{1}] \in R^{m*(k+1)}$, $G = [B_{\_}, e_{2}] \in R^{N_{-neg}{*(k+1)}}$, $Z_{1}=[z_{1}^T, 1]^T \in R^{(k+1)}$, $Z_{2}=[-z_{2}^T, -1]^T \in R^{(k+1)}$, $z_{1}=\mu_{2-}+\kappa_{2}S_{2-}u_{2}$, $z_{1}=\mu_{1}+\kappa_{1}S_{1}u_{1}$, and $||V||$ is equal to $sup_{||u||\leq 1}u^{T}V$.

\begin{figure}
\centerline{\includegraphics[width=1\textwidth,clip=]{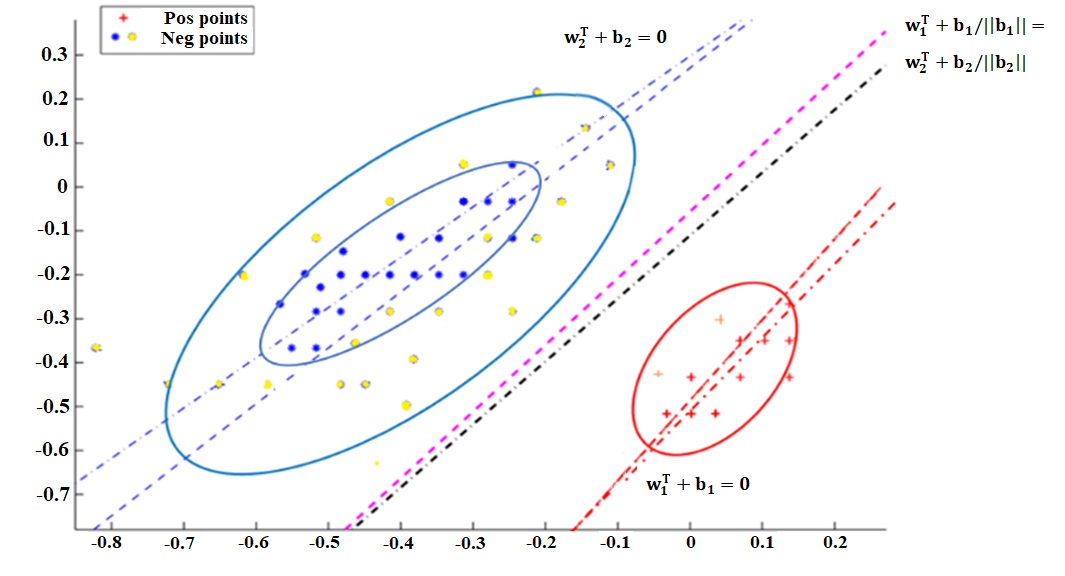}}
\caption{Geometrical interpolation of SOCP-TWSVM and WSOCP-TWSVM.}\label{fig2}      
\end{figure}

So, Fig. (\ref{fig2}) shows the geometrical interpolation of SOCP-TWSVM and WSOCP-TWSVM in a two-dimensional dataset. The blue points are negative samples and yellow points are removed samples
in under-sampling method. The red points are positive samples. The dashed lines represent the
hyperplanes constructed with SOCP-TWSVM. Similarly, the dot-dash lines correspond to the hyperplanes defined by WSOCP-TWSVM. Both methods construct a decision rule that
classifies all training points correctly for the dataset. Although, the decision rules are slightly
different. The method has the advantage that it optimizes both twin hyperplanes in the same
optimization problem, leading to better predictive performance.

\subsection{Nonlinear Weighted SOCP twin support vector machine}\label{subsec3.4}
A kernel-based version can be derived from Eqs. (\ref{eq3}) and (\ref{eq4}) by rewriting weight vector $W \in R^{k}$ as $W = Xs+Mr$ where $M$ is a matrix whose columns are orthogonal to training data points. $S$ and  $r$ are vectors of combining coefficients with the appropriate dimension. $X = [A^{T}, B_{\_}^{T}] \in R^{k*(m+N_{-neg})}$ is the data matrix containing both training patterns. So Kernel-based Twin SOCP-SVM formulation can be written as

\begin{footnotesize}
\begin{equation}\label{eq5}
\begin{matrix}
  \begin{matrix}
    min \\
    s_{1}, b_{1}, \xi_{2}
  \end{matrix} & \frac{1}{2}||K_{1}\bullet s_{1}+e_{1}b_{1}||^2 + \frac{\theta_{1}}{2}(||s_{1}||^2+b_{1}^2)+c_{1}D_{1}\xi_{2} \\
   & \\
  subject~to & - s_{1}^{T}g_{2}-b_{1}\geq 1 -\xi_{2}  + \kappa_{2}||\Lambda_{2}^T s_{1}||.\xi_{2}
\end{matrix}
\end{equation}
\end{footnotesize}

\begin{footnotesize}
\begin{equation}\label{eq6}
\begin{matrix}
  \begin{matrix}
    min \\
    s_{2}, b_{2}, \xi_{1}
  \end{matrix} & \frac{1}{2}||K_{2}\bullet s_{2}+e_{2}b_{2}||^2 + \frac{\theta_{2}}{2}(||s_{2}||^2+b_{2}^2)+c_{2}D_{2}\xi_{1} \\
   & \\
  subject~to &  s_{2}^{T}g_{1}+b_{2}\geq 1 - \xi_{1}+ \kappa_{1}||\Lambda_{1}^T s_{2}||.\xi_{1}
\end{matrix}
\end{equation}
\end{footnotesize}
where using a kernel function, $\Xi_{i}=\Lambda_{i}\Lambda_{i}^T$ for $i=1, 2$, $g_{i}=\frac{1}{m_{1}}\begin{bmatrix} 
K_{1i}e_{i} \\
K_{2i}e_{i}
\end{bmatrix}$ \citep{Nath_2007, Maldonado_2016}.

\section{Experimental Results}\label{sec4}
To show the effectiveness of the proposed WSOCP-TWSVM, we compare it with WSVM, OverSVM, UnderSVM, SMOTESVM, TWSVM and SOCP-TWSVM on 11 standard benchmark datasets from the UCI repository (https://archive.ics.uci.edu/ml/index.php). All methods are implemented in MATLAB R2014a environment.

\subsection{Evaluation Criteria}
The performance of different classifiers is evaluated using confusion matrix. This paper evaluates the performance of proposed methodology for class imbalance using accuracy and AUC. Accuracy of a classifier is estimated by the correct prediction made by the classifier in proportion to total number of prediction.
Sensitivity of a classifier is evaluated by the percentage of positive values that are recognized accurately and also known as true positive rate.
Specificity of a classifier is estimated by the percentage of negative values that are recognized
correctly by the classifier. It is also known as true negative rate. Generally, the G-mean \citep{Chawla_2011} can characterize trade-off between sensitivity and specificity. In our experiments, we also use two common performance measures associated with classifier.
\subsection{Description of datasets and validation procedure}
These benchmark datasets represent a wide range of fields, size and imbalanced ratios. Table \ref{tab1} gives the characteristics of these datasets and minority class for each dataset is shown in the table. The rest of data are classified as the majority one. A grid search is also performed to study the influence of the parameters $\eta$ and $k$ in $KNN$ method for the new approach. In this case, we studied all combinations of the following data, $\eta_{1} = {0.2, 0.4 , 0.6, 0.8}$ and $\eta_{2} = {0.2, 0.4, 0.6, 0.8}$, where $\eta_{1}$ and $\eta_{2}$ are the maximum false positive and false negative error, respectively. The K parameter is selected from the set $\{3,5,10,15\}$, and all other parameters for WSVM, TWSVM, SOCP-TWSVM and our WSOCP-TWSVM is selected from the set $\{10^{-9}, ..., 10^{-6}\}$, also we consider $c_{1}=c_{2}$ and $c_{3}=c_{4}$.
For the above procedure, we employ libsvm \citep{Chang_2011} to be the base classifier of SVM and the SeDuMi MATLAB toolbox as the SOCP-based classifiers \citep{Sturm_1999}.
\subsection{Test using UCI database}
We study the following classification approaches; the Weighted SVM, OverSVM, UnderSVM, SMOTESVM, TWSVM, and SOCP-TWSVM, and WSOCP-TWSVM. For each dataset, seven different classifiers were trained and tested by using nested $10$ cross-validation technique. The accuracy and G-mean of this cross-validation process are averaged for ten runs. The average accuracy of the compared classifiers with linear kernel and the nonlinear kernel is summarized in Tables \ref{tab2} and \ref{tab4}. It can be concluded that  WSOCP-TWSVM has best performance in compared with SOCP-TWSVM in most cases. For instance,  WSOCP-TWSVM in Yeast3 and Heberman datasets is better than other classifiers. Generally, the methods like TWSVM and oversampling have better result for datasets with great imbalanced ratio like   PimaIndian and Ionosphere. Our results on big datasets like Pageblocks infer that that WSOCP-TWSVM possesses a better accuracy in compared with other classifiers. Average G-mean of the compared classifiers with linear kernel and the nonlinear kernel is summarized in Tables \ref{tab3} and \ref{tab5} which indicates the excellence of the performance of WSOCP-TWSVM in compared with SOCP-TWSVM. Tables \ref{tab6} and \ref{tab7} show the training time for these seven classifiers with linear and nonlinear kernel.  The algorithm's execution time naturally increased by running the sampling phase at the beginning of the run. Regarding the execution time in both kernel states, it can be concluded that the sampling phase has a higher overhead time rather than other methods. In particular,  SOCPTWSVM rather than the other classifiers possess the highest execution time for some databases like German and Yeast3. In general, the runtime in the WSOCP-TWSVM algorithm is greater than the SOCPTWSVM method but the accuracy of the new method is higher.

The Friedman test \citep{Friedman_1939} is a non-parametric statistical test that is used to detect differences in treatments across multiple test attempts by considering their ranking. This test is employed here to detect differences by our method. The Friedman test confirms that our strategy is better than other comparable methods in terms of accuracy. For linear cases, the result of the Friedman test is presented in Table \ref{tab8} (a) and (b) and for nonlinear cases, the result of the test is presented in Table \ref{tab8} (c) and (d). The ranking results from the Friedman test show that WSOCP-TWSVM performs better than other methods. Based on results, for linear cases, though the accuracy of the WSOCP-TWSVM is similar to that of  SMOTESVM, the accuracy of WSVM is a little worse than both. Also, the G-mean of WSOCP-TWSVM and SMOTESVM are similar and both have the best performance. The results on nonlinear classifiers have shown that accuracy of the SMOTESVM is a little worse than our WSOCP-TWSVM, the G-mean of SOCP-TWSVM is lower than WSOCP-TWSVM. We can see that our proposed approach obtains the best-imbalanced classification performance than the others in most cases, specifically, it enhances the performance of SOCP-TWSVM.

\section{Conclusion}\label{sec5}
A new method of imbalanced data classification named WSOCP-TWSVM is proposed in the present paper. This method uses the under-sampling procedure for training the dataset and gives the weights for each class. The results of numerical tests performed on datasets show that the proposed methodology is feasible and effective on generalization ability. The WSOCP-TWSVM method is better than the others in the  kernel case. Introducing this method provides opportunities to continue the future works. Our method can be extended to multi-class classifications and used for some practical application. In addition, the employment of some different weight setting methods can improve the performance of the WSOCP-TWSVM method.

\begin{table}[]
\caption{Characteristics of the benchmark datasets.}
\label{tab1}
\begin{tabular}{lllll}
Dataset     & IR     & \#Features & Minority class & Data size \\
Yeast3      & 0.1098 & 8          & ME3            & 1484      \\
Vehicle     & 0.2362 & 18         & VAN            & 946       \\
Transfusion & 0.2380 & 4          & Yes            & 748       \\
Wine        & 0.3315 & 13         & Class 1        & 178       \\
PimaIndian  & 0.3490 & 8          & Diabetes       & 768       \\
Ionosphere  & 0.3590 & 34         & Bad            & 351       \\
Haberman    & 0.2647 & 3          & Died           & 306       \\
German      & 0.3000 & 20         & Bad            & 1000      \\
CMC         & 0.2261 & 9          & Lon-term       & 1473      \\
Yeast4      & 0.0340 & 8          & ME2            & 1484      \\
Wisconsin   & 0.1021 & 9          & Rest           & 683\\ 
Segment &0.1427& 19& Segment& 2308\\
Page-blocks& 0.1021& 10& Rest& 5472
\end{tabular}
\end{table}

\begin{sidewaystable}[]
\caption{The training accuracy of linear classifiers on benchmark datasets}
\label{tab2}
\begin{tabular}{llllllll}
Dataset     & WSVM       & OverSVM    & UnderSVM   & SMOTESVM   & TWSVM      & SOCP-TWSVM   & \textbf{WSOCP-TWSVM} \\
Yeast3      & 90.79$\pm$1.36 & 90.22$\pm$0.15 & 83.78$\pm$0.19 & 91.92$\pm$0.39 & 89.17$\pm$0.41 & 90.56$\pm$3.05  & \textbf{92.35$\pm$3.25} \\
Vehicle     & 95.80$\pm$0.86 & 94.67$\pm$0.10 & 93.83$\pm$0.85 & \textbf{97.03$\pm$0.24} & 96.59$\pm$0.23 & 95.65$\pm$3.98  & 94.23$\pm$1.04 \\
Transfusion & 62.73$\pm$4.28 & 54.86$\pm$2.21 & 55.69$\pm$2.67 & 65.06$\pm$3.73 & 49.43$\pm$2.27 & 59.45$\pm$4.26  & \textbf{65.23$\pm$8.79} \\
Wine        & 92.12$\pm$0.50 & 95.72$\pm$0.55 & 91.20$\pm$0.73 & \textbf{96.04$\pm$0.19} & 94.81$\pm$0.50 & 93.38$\pm$10.10 & 92.92$\pm$7.03 \\
PimaIndian  & 75.64$\pm$0.79 & 75.26$\pm$0.90 & 72.79$\pm$0.53 & 73.01$\pm$0.62 & \textbf{76.10$\pm$0.25} & 71.23$\pm$7.83  & 74.42$\pm$0.36 \\
Ionosphere  & 88.75$\pm$2.29 & \textbf{89.11$\pm$0.67} & 83.62$\pm$0.44 & 73.01$\pm$0.30 & 84.93$\pm$1.21 & 83.73$\pm$4.52  & 84.90$\pm$3.27 \\
Haberman    & 70.68$\pm$0.92 & 64.82$\pm$0.51 & 58.96$\pm$1.74 & 74.77$\pm$0.28 & 75.81$\pm$0.46 & 75.89$\pm$6.67  & \textbf{76.53$\pm$6.72} \\
German      & 72.30$\pm$1.28 & 71.68$\pm$1.58 & 49.99$\pm$0.47 & 69.66$\pm$0.63 & \textbf{74.84$\pm$0.30} & 73.70$\pm$3.12  & 72.90$\pm$6.15 \\
CMC         & 73.46$\pm$0.35 & 71.25$\pm$0.17 & 51.94$\pm$0.78 & 75.83$\pm$0.41 & 74.36$\pm$0.21 & 76.64$\pm$1.17  & \textbf{77.39$\pm$0.26} \\
Yeast4      & 94.29$\pm$0.25 & 84.98$\pm$0.18 & 87.06$\pm$1.73 & \textbf{96.50$\pm$0.26} & 75.10$\pm$0.16 & 90.89$\pm$2.37  & \textbf{92.12$\pm$2.07} \\
Wisconsin   & \textbf{96.72$\pm$0.73} & 93.37$\pm$0.26 & 93.10$\pm$0.24 & 95.92$\pm$0.17 & 95.34$\pm$0.25 & 94.13$\pm$2.75  & 94.02$\pm$9.35\\
\textbf{Segment} & 92.26$\pm$0.21&	92.52$\pm$0.42&	91.26$\pm$0.87&	90.24$\pm$0.32&	91.26$\pm$0.25&	\textbf{95.24$\pm$0.58}&	93.16$\pm$0.45\\
\textbf{Page-blocks} &90.87$\pm$0.25	& 91.24$\pm$2.03&	87.15$\pm$0.13&	90.15$\pm$0.45&	92.54$\pm$0.16&	92.14$\pm$0.18&	\textbf{92.54$\pm$0.36}
\end{tabular}

\end{sidewaystable}

\begin{sidewaystable}[]
\caption{The training G-mean of linear classifiers on benchmark datasets.}
\label{tab3}
\begin{tabular}{lllllllll}
Dataset     & WSVM       & OverSVM    & UnderSVM   & SMOTESVM   & TWSVM      & SOCP-TWSVM   &\textbf{WSOCP-TWSVM} \\
Yeast3      & 78.02$\pm$1.20 & 76.23$\pm$0.15 & 71.75$\pm$1.69 & 84.93$\pm$0.73 & 83.07$\pm$0.96 & 91.48$\pm$4.85  & \textbf{93.23$\pm$3.16}  \\
Vehicle     & 90.38$\pm$0.74 & 91.69$\pm$0.41 & 86.91$\pm$0.80 & \textbf{96.05$\pm$0.27} & 93.74$\pm$0.43 & 89.17$\pm$3.26  & 90.33$\pm$1.62  \\
Transfusion & 58.68$\pm$2.75 & 61.31$\pm$2.72 & 56.97$\pm$2.66 & 62.06$\pm$3.18 & 52.72$\pm$0.36 & 56.82$\pm$4.26  & 63.24$\pm$3.02  \\
Wine        & 76.57$\pm$1.64 & 82.07$\pm$2.56 & 80.20$\pm$2.73 & 84.72$\pm$2.19 & 86.87$\pm$1.36 & 88.02$\pm$8.66  & \textbf{90.52$\pm$0.52}  \\
PimaIndian  & 66.39$\pm$2.16 & 71.24$\pm$0.92 & 68.79$\pm$0.53 & 72.01$\pm$1.51 & 70.43$\pm$2.80 & 63.75$\pm$3.25  & \textbf{71.52$\pm$8.03}  \\
Ionosphere  & 75.59$\pm$1.26 & \textbf{80.54$\pm$0.44} & 62.09$\pm$0.41 & 77.31$\pm$0.93 & 75.67$\pm$0.97 & 76.47$\pm$8.73  & 79.49$\pm$3.82  \\
Haberman    & 60.84$\pm$2.85 & 62.90$\pm$0.67 & 52.15$\pm$3.48 & \textbf{66.92$\pm$0.50} & 52.21$\pm$4.90 & 52.73$\pm$11.51 & 58.96$\pm$4.24  \\
German      & 59.09$\pm$3.62 & 67.09$\pm$3.50 & 61.96$\pm$3.37 & 64.24$\pm$1.83 & 65.47$\pm$2.51 & \textbf{71.95$\pm$4.71}  & 70.03$\pm$5014  \\
CMC         & 52.91$\pm$1.08 & 50.58$\pm$2.17 & 45.56$\pm$1.29 & 60.43$\pm$1.76 & 54.61$\pm$2.22 & 59.83$\pm$5.22  & \textbf{64.91$\pm$5.28}  \\
Yeast4      & 73.48$\pm$2.01 & 82.02$\pm$0.19 & 67.93$\pm$1.97 & 84.35$\pm$0.18 & 18.78$\pm$6.20 & 80.93$\pm$6.79  & \textbf{85.36$\pm$8.85}  \\
Wisconsin   & \textbf{96.49$\pm$0.36} & 91.40$\pm$0.21 & 89.14$\pm$0.24 & 93.54$\pm$0.18 & 92.04$\pm$0.52 & 90.71$\pm$5.72  & 89.86$\pm$7.07\\
\textbf{Segment}&94.12$\pm$0.34&	90.45$\pm$0.15&	84.26$\pm$0.95&	75.15$\pm$0.12&	90.15$\pm$0.89&	\textbf{95.14$\pm$0.23}&	94.15$\pm$0.85\\
\textbf{Page-blocks} &75.25$\pm$0.23&	85.12$\pm$0.31&	70.12$\pm$0.16&	85.12$\pm$0.45&	85.12$\pm$0.13&	90.25$\pm$0.12&	\textbf{91.25$\pm$0.23}
\end{tabular}
\end{sidewaystable}

\begin{sidewaystable}[]
\caption{The training accuracy of nonlinear classifiers on benchmark datasets.}
\label{tab4}
\begin{tabular}{llllllll}
Dataset    & WSVM       & OverSVM    & UnderSVM   & SMOTESVM   & TWSVM      & SOCPTWSV   & \textbf{WSOCP-TWSVM} \\
Yeast3     & 91.39$\pm$1.77 & 92.91$\pm$1.04 & 87.61$\pm$3.42 & \textbf{93.52$\pm$1.49} & 92.31$\pm$0.16 & 91.91$\pm$2.01 & 92.31$\pm$0.78  \\
Vehicle    & 78.54$\pm$0.89 & 80.60$\pm$1.89 & 74.16$\pm$2.56 & 82.80$\pm$1.52 & 79.47$\pm$1.65 & \textbf{94.10$\pm$4.32} & 93.89$\pm$2.43  \\
Transfusio & 73.25$\pm$2.35 & 74.85$\pm$1.38 & 68.59$\pm$1.17 & 75.60$\pm$2.59 & 60.16$\pm$3.29 & \textbf{77.14$\pm$4.76} & 72.31$\pm$3.51  \\
Wine       & 84.28$\pm$2.18 & \textbf{95.73$\pm$3.48} & 77.62$\pm$1.17 & 92.51$\pm$2.64 & 92.15$\pm$2.07 & 92.16$\pm$5.70 & 92.19$\pm$3.67  \\
PimaIndia  & 62.79$\pm$1.61 & 65.10$\pm$0.37 & 64.42$\pm$2.28 & 73.42$\pm$1.78 & 72.23$\pm$1.72 & 73.43$\pm$3.24 & \textbf{73.53$\pm$4.69}  \\
Ionosphere & 93.67$\pm$1.59 & 91.63$\pm$2.64 & 87.45$\pm$2.36 & \textbf{94.32$\pm$1.71} & 90.30$\pm$0.47 & 92.90$\pm$3.26 & 90.31$\pm$5.35  \\
Haberman   & 73.55$\pm$3.14 & 74.24$\pm$1.15 & 58.96$\pm$2.68 & 72.77$\pm$2.41 & 69.39$\pm$1.56 & 75.83$\pm$6.28 & \textbf{76.03$\pm$5.33}  \\
German     & 71.24$\pm$3.01 & 68.45$\pm$0.39 & 60.43$\pm$1.58 & 71.53$\pm$0.29 & 70.47$\pm$1.86 & 70.00$\pm$3.43 & \textbf{72.50$\pm$4.37}  \\
CMC        & 62.25$\pm$2.07 & 68.13$\pm$1.65 & 60.27$\pm$1.89 & 71.74$\pm$2.20 & 73.62$\pm$1.55 & \textbf{78.34$\pm$2.34} & 78.12$\pm$7.14  \\
Yeast4     & 92.63$\pm$1.39 & 94.07$\pm$2.86 & 92.49$\pm$1.73 & 95.48$\pm$1.82 & 92.16$\pm$2.79 & 94.55$\pm$0.20 & \textbf{95.67$\pm$3.02}  \\
Wisconsin  & 93.65$\pm$2.25 & 94.26$\pm$0.86 & 91.28$\pm$2.58 & 96.16$\pm$0.81 & 96.02$\pm$0.72 & 95.85$\pm$1.70 & \textbf{96.56$\pm$2.28}\\
\textbf{Segment}&78.15$\pm$0.25&	89.15$\pm$0.45&	65.14$\pm$0.45&	62.24$\pm$0.15&	89.12$\pm$0.14&	\textbf{90.15$\pm$2.13}&	89.84$\pm$2.03\\
\textbf{Page-blocks} &\textbf{90.15$\pm$0.23}&	89.14$\pm$2.01&	85.16$\pm$0.45&	89.15$\pm$1.02&	89.14$\pm$0.17&	90.01$\pm$2.01&	90.05$\pm$0.30
\end{tabular}
\end{sidewaystable}

\begin{sidewaystable}[]
\caption{The training G-mean of nonlinear classifiers on benchmark datasets.}
\label{tab5}
\begin{tabular}{llllllll}
Dataset    & WSVM       & OverSVM    & UnderSVM   & SMOTESVM   & TWSVM      & SOCP-TWSVM   &  \textbf{WSOCP-TWSVM}\\
Yeast3     & 81.05$\pm$1.46 & 80.73$\pm$0.45 & 74.14$\pm$1.25 & 85.32$\pm$0.83 & 78.68$\pm$2.57 & \textbf{91.99$\pm$2.01}  & 91.60$\pm$3.37  \\
Vehicle    & 81.05$\pm$1.39 & 84.31$\pm$0.42 & 69.41$\pm$2.44 & 90.61$\pm$1.29 & 63.08$\pm$3.30 & 90.56$\pm$7.26  & \textbf{91.99$\pm$4.56}  \\
Transfusio & 57.55$\pm$1.83 & 50.89$\pm$2.89 & 43.97$\pm$1.50 & 60.52$\pm$2.04 & 51.90$\pm$1.21 & 61.21$\pm$6.99  & \textbf{63.96$\pm$5.14}  \\
Wine       & 7.52$\pm$18.94 & 71.93$\pm$4.62 & 0.00$\pm$0.00  & 67.44$\pm$2.81 & 60.87$\pm$2.68 & 91.73$\pm$5.75  & \textbf{92.39$\pm$7.57}  \\
PimaIndia  & 48.63$\pm$2.73 & 54.89$\pm$3.11 & 19.74$\pm$8.51 & 58.42$\pm$3.20 & 57.16$\pm$2.68 & 70.54$\pm$3.31  & \textbf{71.04$\pm$4.12}  \\
Ionosphere & 93.93$\pm$1.60 & \textbf{94.07$\pm$2.56} & 72.44$\pm$2.96 & 93.35$\pm$2.95 & 90.73$\pm$3.14 & 93.67$\pm$3.13  & 93.84$\pm$8.11  \\
Haberman   & 35.52$\pm$1.47 & 43.78$\pm$11.1 & 29.26$\pm$2.18 & 51.61$\pm$1.58 & 41.37$\pm$2.36 & 60.77$\pm$10.24 & \textbf{63.00$\pm$10.53} \\
German     & 59.25$\pm$0.73 & 58.79$\pm$1.31 & 0.00$\pm$0.00  & 65.12$\pm$0.47 & 54.12$\pm$3.25 & 59.74$\pm$7.17  & \textbf{65.23$\pm$7.33}  \\
CMC        & 64.04$\pm$7.38 & 62.23$\pm$2.98 & 54.34$\pm$2.00 & 41.05$\pm$5.40 & 53.64$\pm$2.19 & 66.18$\pm$3.91  & \textbf{66.41$\pm$5.19}  \\
Yeast4     & 31.13$\pm$3.01 & 51.24$\pm$1.05 & 40.17$\pm$4.15 & 48.08$\pm$1.63 & 46.66$\pm$2.54 & 60.36$\pm$8.91  & \textbf{62.64$\pm$10.21} \\
Wisconsin  & 75.98$\pm$0.53 & 84.26$\pm$8.29 & 90.48$\pm$2.51 & 94.44$\pm$1.32 & 93.85$\pm$3.20 & 95.22$\pm$2.04  & \textbf{96.85$\pm$1.90}\\
\textbf{Segment}&59.15$\pm$0.23&	60.15$\pm$0.58&	65.15$\pm$0.26&	62.23$\pm$0.68&	80.95$\pm$2.01&	\textbf{90.26$\pm$0.12}&	89.12$\pm$3.01\\
\textbf{Page-blocks}&76.25$\pm$2.09&	85.12$\pm$0.10&	90.12$\pm$0.85&	91.25$\pm$2.05&	94.15$\pm$0.10&	93.12$\pm$0.15&	\textbf{94.16$\pm$0.32}
\end{tabular}
\end{sidewaystable}

\begin{sidewaystable}[]
\caption{The training time of linear classifiers on benchmark datasets.}
\label{tab6}
\begin{tabular}{llllllll}
Dataset    & WSVM       & OverSVM    & UnderSVM   & SMOTESVM   & TWSVM      & SOCP-TWSVM   &  \textbf{WSOCP-TWSVM}\\
Yeast3     & 0.3925&	0.1389&	0.0523	&0.0255&	0.9262&	0.6223&	0.8526  \\
Vehicle    & 0.5615	&2.1526	&0.8342	&0.9523	&1.8954	&0.4562	&0.9856 \\
Transfusio & 38.9526	&46.2535	&16.2589	&26.4590	&0.8962	&0.4895	&68.2568  \\
Wine       & 0.9523	&0.3214	&0.2561	&0.4151	&0.0059	&0.1452	&0.6231\\
PimaIndia  & 2.0311	&8.1645	&2.0030	&6.0310	&0.0215	&0.0279	&0.08536  \\
Ionosphere & 0.0521	&0.0295	&0.0152	&0.2560	&0.0310	&0.2613	&0.0361  \\
Haberman   & 0.1652	&0.1025	&0.0214	&0.1389	&0.0321	&0.2140	&0.6231 \\
German     & 29.4510	&35.1500	&15.2301	&23.1547	&0.8925	&0.5216	&0.6954  \\
CMC        & 3.2514	&1.3259	&15.1614&	16.2400&	0.7316	&1.1632	&2.0311 \\
Yeast4     & 0.2798	&0.1624	&0.0921	&0.1232	&0.1456	&0.1315	&0.4510 \\
Wisconsin  & 0.2151	&0.1361	&0.9526	&0.2920	&0.0258	&0.3621	&0.8925\\
\textbf{Segment}&1.2561	&0.8745	&0.9325	&0.6214	&0.6258	&0.7546	&0.9847\\
\textbf{Page-blocks}&94.1561	&195.1232	&39.1621	&354.1212	&45.6251	&23.1514	&38.7916
\end{tabular}
\end{sidewaystable}

\begin{sidewaystable}[]
\caption{The training time of nonlinear classifiers on benchmark datasets.}
\label{tab7}
\begin{tabular}{llllllll}
Dataset    & WSVM       & OverSVM    & UnderSVM   & SMOTESVM   & TWSVM      & SOCP-TWSVM   &  \textbf{WSOCP-TWSVM}\\
Yeast3     & 0.5961	0.1988	&0.0958	&0.1523	&0.0981	&0.02359&	0.1650  \\
Vehicle    & 0.8945	& 0.3012	&0.0952	&0.2135	&0.6254&	0.4512	&0.6895 \\
Transfusio & 0.6250	&0.1523	&0.0258	&0.0987&	0.0984&	0.9890	&1.0230  \\
Wine       & 0.0231	&0.1325	&0.0102	&0.0189	&0.0236	&0.0166	&0.3250 \\
PimaIndia  & 0.0625	&0.0562	&0.0154	&0.0231	&0.1541	&0.0950	&0.1621  \\
Ionosphere & 0.0352&	0.2315&	0.0165	&0.0152	&0.0925&	0.0239	&0.1451 \\
Haberman   & 0.0451&	0.1298&	0.1451&	0.1241&	0.1648&	0.0252	&0.2145\\
German     & 3.0261	&3.1203	&3.1285	&0.0231&	0.2378&	0.0234	&0.2611 \\
CMC        & 2.3714	&0.2315	&0.1458	&0.1485	&0.1547&	0.1898	&0.5210  \\
Yeast4     &0.2154&	0.0591&	0.0721&	0.0915&	0.1524&	0.0699	&0.5921 \\
Wisconsin  & 0.1584	&0.7925	&0.7891	&0.0214	&0.0599&	0.9851	&1.2561\\
\textbf{Segment}&1.8548	&0.9550	&0.0214	&0.0528&	0.0951&	0.9512	&1.0320\\
\textbf{Page-blocks}&31.0252	&7.8912	&0.98521	&6.2511	&46.0259	&21.0985	&36.0695
\end{tabular}
\end{sidewaystable}

\begin{sidewaystable}[]
\caption{Friedman test rank of accuracy and G-mean of linear and nonlinear classifiers. (a) acc-lin (b) G-mean-lin (c)acc-non-lin (d) G-mean-nonlin.}
\label{tab8}
\begin{tabular}{lllllllll}
            & Mean rank &  &Mean rank  &  &Mean rank  &  &Mean rank  &  \\
WSVM        & 4.73      & WSVM & 3.18 &WSVM  & 3.00 &WSVM  & 3.09 \\
OverSVM     & 3.36      &OverSVM   & 4.45 &OverSVM   & 4.18 &OverSVM   & 3.82   \\
UnderSVM    & 1.55      &UnderSVM  & 1.64 &UnderSVM  &  1.27&UnderSVM  & 1.55  \\
SMOTESVM    & 4.91      &SMOTESVM  & 5.64 &SMOTESVM  & 5.55 &SMOTESVM  & 4.45 \\
TWSVM       & 4.36      &TWSVM     &  3.55&TWSVM     & 3.14 &TWSVM     &    2.64  \\
SOCP-TWSVM   & 4.18      &SOCP-TWSVM  & 3.91 &SOCP-TWSVM  &5.18  &SOCP-TWSVM  & 5.73  \\
\textbf{WSOCP-TWSVM} & \textbf{4.93}      & \textbf{WSOCP-TWSVM} & \textbf{5.66} &\textbf{WSOCP-TWSVM}  & \textbf{5.68} &\textbf{WSOCP-TWSVM}  &   \textbf{6.98} \\
(a)        &           &  (b)&  &(c)  &  &(d)  &   
\end{tabular}
\end{sidewaystable}
\bibliography{library}
\end{document}